\documentclass{article} %
\usepackage{main,times}

\usepackage{amsmath,amsfonts,bm}

\def\eqref#1{equation~\ref{#1}}

\def\1{\bm{1}}

\def\va{{\bm{a}}}

\def\vi{{\bm{i}}}

\def\vp{{\bm{p}}}
\def\vq{{\bm{q}}}

\def\vv{{\bm{v}}}

\def\vx{{\bm{x}}}

\def\mI{{\bm{I}}}

\def\mT{{\bm{T}}}

\DeclareMathAlphabet{\mathsfit}{\encodingdefault}{\sfdefault}{m}{sl}
\SetMathAlphabet{\mathsfit}{bold}{\encodingdefault}{\sfdefault}{bx}{n}

\DeclareMathOperator*{\argmin}{arg\,min}

\newcommand{\ourmethod}{{\it PF-LRM}}

\definecolor{citeblue}{RGB}{48,111,186}
\usepackage[pagebackref=false,breaklinks=true,colorlinks=true,citecolor=citeblue,bookmarks=false]{hyperref}
\usepackage{cleveref}  %

\usepackage{hyperref}
\usepackage{url}
\usepackage{graphicx}
\usepackage{capt-of}
\usepackage{multirow}

\title{PF-LRM: \textbf{P}ose-\textbf{F}ree \textbf{L}arge \textbf{R}econstruction \textbf{M}odel for Joint Pose and Shape Prediction}

\author{Peng Wang\thanks{This work is done while the author is an intern at Adobe Research.} \\
Adobe Research \& HKU \\
\texttt{totoro97@outlook.com} \\
\And
Hao Tan \\
Adobe Research \\
\texttt{hatan@adobe.com} \\
\And
Sai Bi \\
Adobe Research \\
\texttt{sbi@adobe.com} \\
\And
Yinghao Xu$^*$ \\
Adobe Research \& Stanford \\
\texttt{yhxu@stanford.edu} \\
\And
Fujun Luan \\
Adobe Research \\
\texttt{fluan@adobe.com} \\
\And
Kalyan Sunkavalli \\
Adobe Research \\
\texttt{sunkaval@adobe.com} \\
\And
Wenping Wang \\
Texas A\&M University \\
\texttt{wenping@tamu.edu} \\
\And
Zexiang Xu \\
Adobe Research \\
\texttt{zexu@adobe.com} \\
\And
Kai Zhang \\
Adobe Research \\
\texttt{kaiz@adobe.com} \\
}

\iclrfinalcopy %
\begin{document}

\maketitle

\begin{abstract}
We propose a \textbf{P}ose-\textbf{F}ree \textbf{L}arge \textbf{R}econstruction \textbf{M}odel (\ourmethod{}) for reconstructing a 3D object from a few \textit{unposed} images even with little visual overlap, while simultaneously estimating the relative camera poses in $\sim$1.3 seconds on a single A100 GPU. 
\ourmethod{} is a highly scalable method utilizing the self-attention blocks to exchange information between 3D object tokens and 2D image tokens; we predict a coarse point cloud for each view, and then use a differentiable Perspective-n-Point (PnP) %
solver to obtain camera poses. 
When trained on a huge amount of multi-view posed data of $\sim$1M objects, \ourmethod{} shows strong cross-dataset generalization ability, and outperforms baseline methods by a large margin in terms of pose prediction accuracy and 3D reconstruction quality on various unseen evaluation datasets. 
We also demonstrate our model's applicability in downstream text/image-to-3D task with fast feed-forward inference.
Our project website is at: \url{https://totoro97.github.io/pf-lrm}.

\end{abstract}

\section{Introduction}
\label{sec:introduction}

3D reconstruction is a classical computer vision problem, with applications spanning imaging, perception, and computer graphics. While both traditional photogrammetry~\citep{barnes2009patchmatch, schoenberger2016mvs, furukawa2009accurate} and modern neural reconstruction methods~\citep{mildenhall2020nerf, wang2021neus, yariv2021volume}
have made significant progress in high-fidelity geometry and appearance reconstruction, they rely on having images with calibrated camera poses as input. These poses are typically computed using a Structure-from-Motion (SfM) solver~\citep{schonberger2016structure, snavely2006photo}.

SfM assumes dense viewpoints of the scene where input images have sufficient overlap and matching image features. This is not applicable in many cases, e.g., e-commerce applications, consumer capture scenarios, and dynamic scene reconstruction problems, 
where adding more views incurs a higher cost and thus the captured views tend to be \emph{sparse} and have a \emph{wide baseline}  
(i.e., share little overlap). In such circumstances, SfM solvers become unreliable and tend to fail. As a result, (neural) reconstruction methods, including sparse methods~\citep{niemeyer2022regnerf,kangle2021dsnerf,long2022sparseneus,zhou2023sparsefusion,wang2023sparsenerf} %
that require accurate camera poses, cannot be reliably used for such applications.

\begin{figure}
    \vspace{-5pt}
    \includegraphics[width=1.\textwidth]{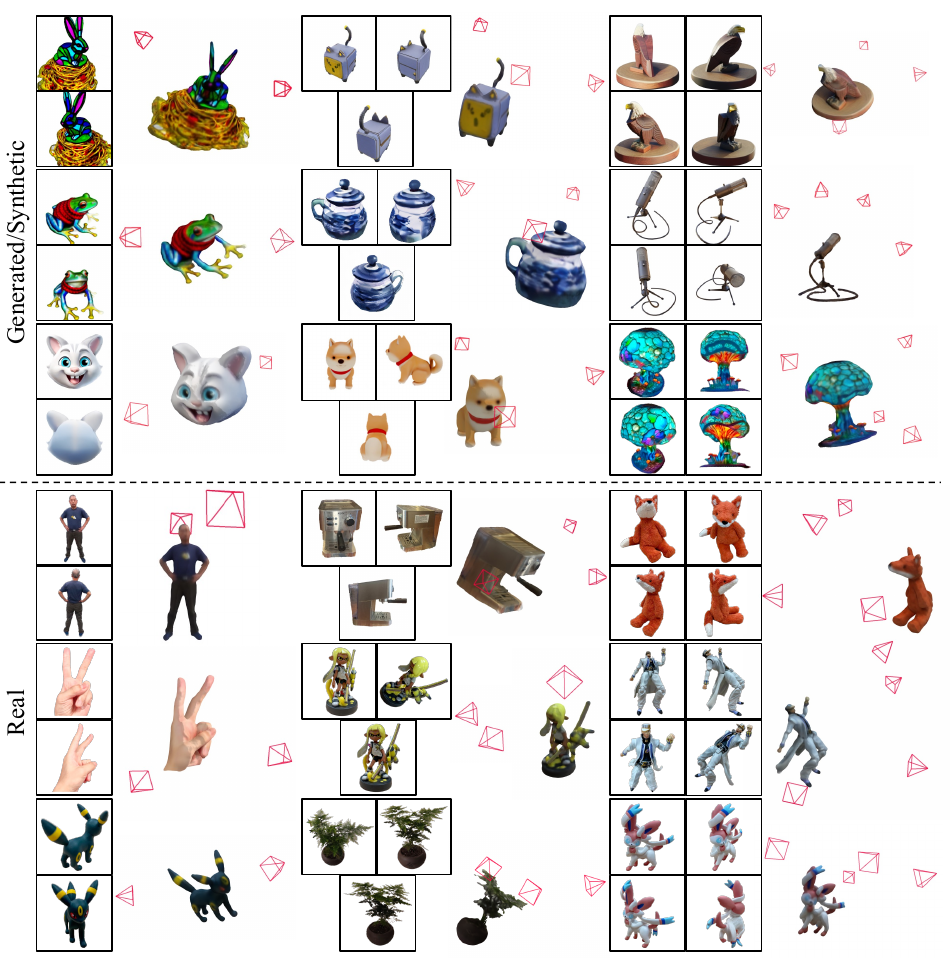}
    \captionof{figure}{
    (Top block) To demonstrate our model's generalizability to unseen in-the-wild images, 
    we take 2-4 unposed images from prior/concurrent 3D-aware generation work, and use our \ourmethod{} to jointly 
    reconstruct the NeRF and  
    estimate relative poses in a feed-forward manner.
    (Bottom block) we also show our model's generalizability on real captures.
    Sources of generated/synthetic images: Column 1 (top-to-bottom),  Magic3D~\citep{lin2023magic3d}, DreamFusion~\citep{poole2022dreamfusion}, Wonder3D~\citep{long2023wonder3d}; Column 2 (top-to-bottom), Zero-1-to-3~\citep{liu2023zero1to3}, SyncDreamer~\citep{liu2023syncdreamer}, Consistent-1-to-3~\citep{ye2023consistent}; Column 3 (top-to-bottom), MVDream~\citep{shi2023MVDream}, NeRF~\citep{mildenhall2020nerf}, Zero123++~\citep{shi2023zero123plus}.
    Source of real images: Row 1 column 1, HuMMan Dataset~\citep{cai2022humman}; Row 1 column 2, RelPose++~\citep{lin2023relposepp}; Others, our phone captures.
    }
    
    \label{fig:teaser}
    \vspace{10pt}
\end{figure}

In this work, we present \ourmethod, a category-agnostic method for jointly predicting both camera poses and object shape and appearance 
(represented using a triplane NeRF~\citep{chan2022efficient, peng2020convolutional}). %
As shown in Fig.~\ref{fig:teaser}, our approach can robustly reconstruct accurate poses and realistic 3D objects using as few as 2--4 sparse input images from diverse input sources.  
The core of our approach is a novel scalable single-stream transformer model (see Fig.~\ref{fig:method}) 
that computes self-attention over the union of the two token sets: the set of 2D multi-view image tokens and the set of 3D triplane NeRF tokens, 
allowing for comprehensive information exchange across all 2D and 3D tokens. We use the final NeRF 
tokens, contextualized by 2D images, to represent a triplane NeRF and supervise it by a novel view rendering loss. 
On the other hand, we use the final image patch tokens contextualized by NeRF tokens for predicting the coarse point clouds used to solve per-view camera poses.

Unlike previous methods that regress pose parameters from images directly, we estimate the 3D object points corresponding to 2D patch centers from their individual patch tokens (contextualized by NeRF tokens). These points are supervised by the NeRF geometry in an online manner during training
and enable accurate pose estimation using a differentiable Perspective-n-Point (PnP) solver~\citep{chen2022epro}. 
In essence, we transform the task from per-view pose prediction into per-patch 3D 
surface point prediction, which is more suitable for our single-stream transformer that's designed for token-wise operations, leading to more accurate results than direct pose regression.

\ourmethod\ is a large transformer model with 
$\sim$590 million
parameters trained on large-scale multi-view posed renderings from Objaverse ~\citep{objaverse} and real-world captures from MVImgNet~\citep{yu2023mvimgnet} that cover $\sim$1 million objects in total,  without direct 3D supervision. Despite being trained under the setting of 4 input views, %
it generalizes well to unseen datasets and can handle a variable number of 2--4 unposed input images during test time (see Fig.~\ref{fig:teaser}),
achieving state-of-the-art results for both pose estimation and novel view synthesis in the case of very sparse inputs, outperforming baseline methods~\citep{jiang2022forge,lin2023relposepp} by a large margin. 
We also showcase some potential downstream applications of our model, e.g., text/image-to-3D, in Fig.~\ref{fig:teaser}.

\section{Related work}
\label{sec:relatedwork}
{\bf NeRF from sparse posed images. } The original NeRF technique~\citep{mildenhall2020nerf} required hundreds of posed images for accurate reconstruction. Recent research on sparse-view NeRF reconstruction has proposed either regularization strategies~\citep{wang2023sparsenerf, niemeyer2022regnerf, yang2023freenerf,kim2022infonerf} or learning priors from extensive datasets~\citep{yu2021pixelnerf, chen2021mvsnerf, long2022sparseneus, ren2023volrecon, zhou2023sparsefusion, irshad2023neo360, li2023instant3d, xu2023dmv3d}.
These approaches still assume precise camera poses for every input image; however determining camera poses given such sparse-view images is non-trivial and off-the-shelf camera estimation pipelines~\citep{schonberger2016structure, snavely2006photo} tend to fail. In contrast, our method efficiently reconstructs a triplane NeRF \citep{chan2022efficient,chen2022tensorf, peng2020convolutional} from sparse views without any camera pose inputs; moreover, our method is capable of recovering the unknown relative camera poses during inference time. %

{\bf Structure from Motion.} %
Structure-from-Motion (SfM) techniques~\citep{schonberger2016structure, snavely2006photo, mohr1995relative} find 2D feature matches across views, and then solve for camera poses and sparse 3D scene structure from these 2D correspondences at the same time. These methods work pretty well in the presence of sufficient visual overlap between nearby views and adequate discriminative features, leading to accurate camera estimation. However, when the input views are extremely sparse, for instance, when there are only 4 images looking from the front-, left-, right-, back- side of an object, it becomes very challenging to match features across views due to the lack of sufficient overlap, even with modern learning-based feature extractors~\citep{detone2018superpoint, dusmanu2019d2, revaud2019r2d2} and matchers~~\citep{sarlin2020superglue,sarlin2019hloc,liu2021learnable}. In contrast, our method relies on the powerful learnt shape prior from a large amount of data to successfully register the cameras in these challenging scenarios.

{\bf Neural pose prediction from RGB images.} 
A series of methods~\citep{lin2023relposepp, rockwell20228, cai2021extreme} have sought to address this issue by directly regressing camera poses through network predictions.
Notably, these methods do not incorporate 3D shape information during the camera pose prediction process. 
We demonstrate that jointly reasoning about camera pose and 3D shape leads to significant improvement over these previous methods that only regress the camera pose.  SparsePose~\citep{sinha2023sparsepose}, FORGE~\citep{jiang2022forge} and FvOR~\citep{yang2022fvor} implement a two-stage prediction pipeline, initially inferring coarse camera poses 
 and coarse shapes by neural networks and then refining these pose predictions  (through further network evaluations~\citep{sinha2023sparsepose} or per-object optimizations~\citep{jiang2022forge, yang2022fvor}) jointly with 3D structures. Our method employs a single-stage inference pipeline to recover both camera poses and 3D NeRF reconstructions at the same time. To predict camera poses, we opt not to regress them directly as in prior work. Instead, we predict a coarse point cloud in the scene coordinate frame~\citep{shotton2013scene} for each view from their image patch tokens; these points, along with image patch centers, establish a set of 3D-2D correspondences, and allow us to solve for poses using a differentiable PnP solver~\citep{chen2022epro, brachmann2017dsac}. This is in contrast to solving poses from frame-to-frame scene flows (3D-3D correspondences) used by FlowCam~\citep{smith2023flowcam}, and better suits the sparse view inputs with little overlap. Moreover, our backbone model is a simple transformer-based model that is highly scalable; hence it can be trained on massive multi-view posed data of diverse and general objects to gain superior robustness and generalization. This distinguishes us from the virtual correspondence work~\citep{Ma_2022_CVPR} that's  designed specifically for human images.

\newcommand{\Img}{{\mI}}
\newcommand{\Pose}{{y}}
\newcommand{\Tripl}{\mT}

\newcommand{\VEncoding}{\vv}
\newcommand{\IEncoding}{\vi}
\newcommand{\TokenD}{{D}}
\newcommand{\Color}{{C}}
\newcommand{\Pos}{{\vx}}
\newcommand{\Trans}{{\tau}}
\newcommand{\Dens}{{\sigma}}
\newcommand{\Rad}{{c}}
\newcommand{\Loss}{{L}}

\newcommand{\ImgToken}{\va}
\newcommand{\ImgPoint}{\vp}
\newcommand{\ImgPixel}{\vq}
\newcommand{\Imgalpha}{\alpha}
\newcommand{\ImgConf}{w}
\newcommand{\ProjLoss}{\xi}

\section{Method}
\label{sec:method}

\begin{figure} \label{fig:method}
\includegraphics[width=1\textwidth]{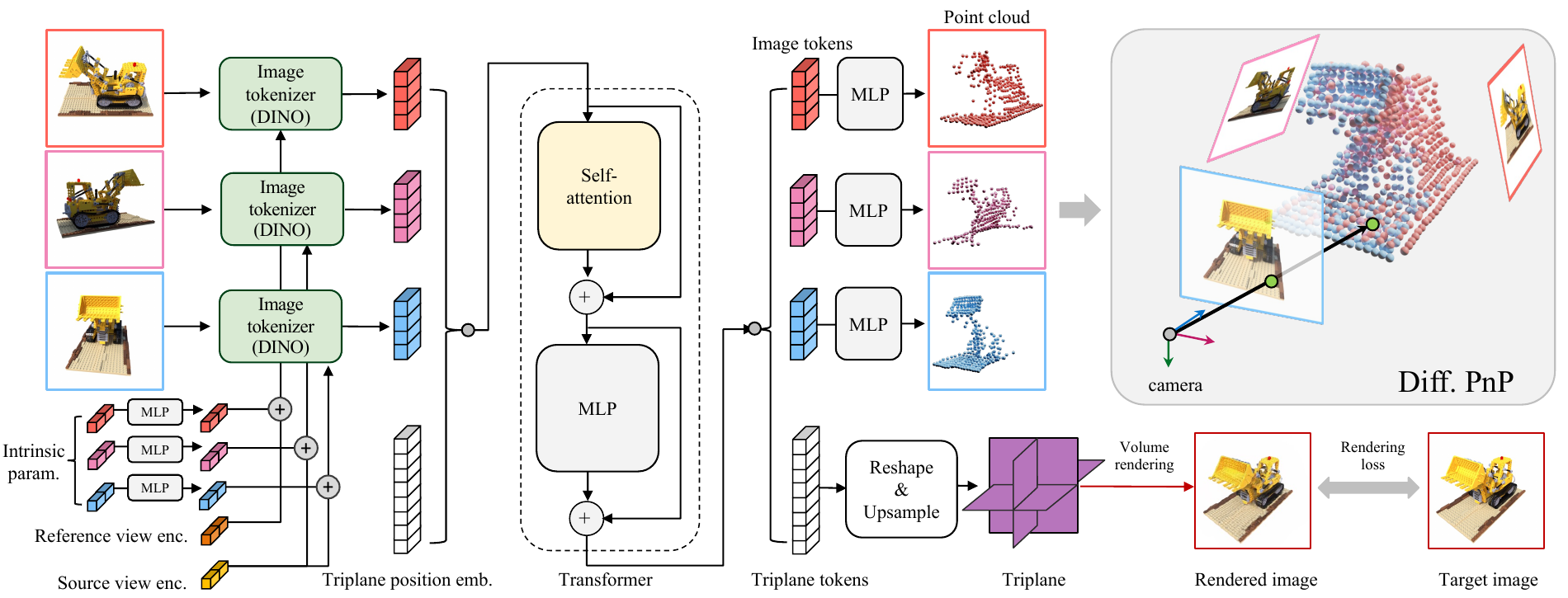}
\vspace{-1.5\baselineskip}
\caption{
\textbf{Overview of our pipeline.} Given unposed sparse input images, we use a large transformer model to reconstruct a triplane NeRF while simultaneously estimating the relative camera poses of all source views with respect to the reference one. 
During training, the triplane tokens are supervised with a rendering loss at novel viewpoints using ground-truth camera poses.  
For camera registration, instead of directly regressing the camera poses, 
we map the image tokens to a coarse 3D geometry in the form of a point cloud (top right),
where we predict a 3D point from each patch token corresponding to the patch center. We then use a differentiable PnP solver to obtain the camera poses from these predicted 3D-2D correspondences  (Sec.~\ref{sec:pose}).  
}
\end{figure}

Given a set of $N$ images $\{\Img_i|i=1,..,N\}$ with unknown camera poses capturing a 3D object, our goal is to reconstruct the object's 3D model and estimate the pose of each image. In particular, we designate one input image (i.e., $\Img_1$)  as a \emph{reference} view, and predict a 3D triplane NeRF and camera poses of other images relative to the reference view. This is expressed by
\begin{equation}
\Tripl, \Pose_2, ..., \Pose_N = \text{PF-LRM}(\Img_1,...,\Img_N),
\end{equation}
where $\Tripl$ is the triplane NeRF defined in the coordinate frame of the reference view 1 and $\Pose_2$,...,$\Pose_N$ are the predicted camera poses of view $2,\ldots,N$ relative to view 1.

We achieve this using a transformer model as illustrated in Fig.~\ref{fig:method}.
Specifically, we tokenize both input images and 
a triplane NeRF, and apply a single-stream multimodal transformer~\citep{chen2020uniter, li2019visualbert} to process the concatenation of NeRF tokens and image patch tokens with self-attention layers (Sec.~\ref{sec:transformer}).
The output
NeRF tokens represent a triplane NeRF for neural rendering, modeling object's geometry and appearance (Sec.~\ref{sec:rendering}), and the %
output image patch tokens are used to 
estimate per-view coarse point cloud for pose estimation with a differentiable PnP solver (Sec.~\ref{sec:pose}).

\subsection{Single-stream Transformer}
\label{sec:transformer}

\noindent \textbf{Image tokenization, view encoding, intrinsics conditioning}. We use the pretrained DINO~\citep{caron2021emerging} Vision Transformer~\citep{dosovitskiy2020image} to tokenize our input images. 
We specifically take DINO-ViT-B/16 with a patch size of $16 \times 16$ and a 12-layer transformer of width $\TokenD=768$. 
Each input image of resolution $H\times W$ is tokenized into $M=H/16 \times W/16$ tokens. 

To distinguish reference image tokens from source image tokens, we use two additional learnable 768-dimensional features, $\VEncoding_r$ and $\VEncoding_s$, as view encoding vectors -- one $\VEncoding_r$ for the reference view ($i=1$) and another $\VEncoding_s$ for all other source views ($i=2,..,N$). 
These view encoding vectors 
allow our model to perform shape reconstruction and pose estimation relative to the reference view. 
In addition, to make our model aware of input cameras' intrinsics, we use a shared MLP to map each view's intrinsics $\left[ f_x, f_y, c_x, c_y\right]$ to a intrinsics conditioning vector $\IEncoding \in \mathbf{R}^{768}$; hence we have $\IEncoding_r, \IEncoding_i, i=2,...,N$ for reference and source views, respectively. We then pass the addition of each view's view encoding and intrinsics conditioning vectors to the newly-added adaptive layer norm block inside each transformer block (self-attention + MLP), following prior work~\citep{hong2023lrm, Peebles2022DiT, huang2017arbitrary}.

\noindent \textbf{Triplane tokenization and position embedding}.
We tokenize a triplane $\Tripl$  of shape $3 \! \times \! H_T \! \times \! W_T \times \! D_T$ into $3 \! \times \! H_T \! \times \! W_T$ tokens, where $H_T, W_T, D_T$ denote triplane height, width and channel, respectively. We additionally learn a triplane position embedding $\Tripl_{pos}$ consisting of $3 \! \times \! H_T \! \times \! W_T$ position markers for triplane tokens; they are mapped to the target triplane tokens by a transformer model sourcing information from input image tokens.

\noindent \textbf{Single-stream transformer}.
The full process of this single-stream transformer can be written as
\begin{equation}
\Tripl, \{\ImgToken_{i,j}| i=1,..,N;j=1,...,M\} = \text{PF-LRM}(\Tripl_{pos}, \Img_1,...,\Img_N, \VEncoding_r, \VEncoding_s).
\end{equation}

Here $\ImgToken_{i,j}$ represents the token of the $j^\text{th}$ patch at view $i$,
and \text{PF-LRM} is 
a sequence of transformer layers.
Each transformer layer 
is composed of a self-attention layer and a multi-layer perceptron layer (MLP), where both use residual connections. We simply concatenate the image tokens and the triplane tokens as Transformer's input as shown in Fig.~\ref{fig:method}.
The output triplane tokens $\Tripl$ and image tokens $\ImgToken_{i,j}$ are used for volumetric NeRF rendering and per-view pose prediction, which we will discuss later.
Our model design is inspired by LRM~\citep{hong2023lrm} and its follow-ups~\citep{li2023instant3d, xu2023dmv3d}, but are different and has its own unique philosophy in that we adopt a single-stream architecture where information exchange is mutual between image tokens and NeRF tokens due to that we predict both a coherent NeRF and per-view coarse geometry used for camera estimation  (detailed later in Sec.~\ref{sec:pose}), while prior work adopts an encoder-decoder design where NeRF tokens source unidirectional information from image tokens using cross-attention layers.

\subsection{NeRF Supervision via Differentiable Volume rendering}
\label{sec:rendering}
To supervise the learning of shape and appearance, 
we use neural differentiable volume rendering to render images at novel viewpoints from the triplane NeRF, as done in \citep{mildenhall2020nerf,chan2022efficient}. This process is expressed by
\begin{equation}
\begin{gathered}
    \Color =\sum_{k=1}^{K} \Trans_{k-1}(1-\exp (-\Dens_{k} \delta_{k})) \Rad_{k}, \quad \Trans_{k}=\exp(-\sum_{k'=1}^{k} \Dens_{k'} \delta_{k'}), 
    \quad \Dens_k,\Rad_k = \text{MLP}_{\Tripl}( \Tripl(\Pos_k)).
\end{gathered} 
\label{eqn:nerfrendering}
\end{equation}
Here, $\Color$ is the rendered RGB pixel color, $\Dens_k$ and $\Rad_k$ are volume density and color decoded from the triplane NeRF $\Tripl$ at the 3D location $\Pos_k$ on the marching ray through the pixel, and $\Trans_{k}$ ($\Trans_{0}$ is defined to be 1) and $\delta_k$ are the volume transmittance and step size; $\Tripl(\Pos_k)$ represents the features that are bilinearly sampled and concatenated from the triplane at $\Pos_k$, and we apply an MLP network $\text{MLP}_{\Tripl}$ to decode the density and color used in volume rendering. 

We supervise our NeRF reconstruction with L2 and VGG-based LPIPS~\citep{zhang2018perceptual} rendering loss:
\begin{equation}\label{eq:rendering_loss}
    \Loss_\Color = \gamma_{\Color}'\ \|\Color-\Color_{gt}\|^2+\gamma_{\Color}''\ \Loss_{lpips}(\Color, \Color_{gt}),
\end{equation}

where $\Color_{gt}$ is the ground-truth pixel color, and $\gamma_{\Color}',\gamma_{\Color}''$ are loss weights. In practice, we render crops of size $h\times w$ for each view to compute the rendering loss $\Loss_\Color$, and divide the L2 loss with $h\times w$. 

\subsection{Pose prediction via Differentiable PnP Solver}
\label{sec:pose}
We estimate relative camera poses from the per-view image patch tokens contextualized by the NeRF tokens.
Note that a straightforward solution is to directly regress camera pose parameters from the image tokens  using an MLP decoder and supervise the poses with the ground truth;
however, such a na\"ive solution lacks 3D inductive biases and, in our experiments (See Tab.~\ref{tab:ablation_pose}), 
often leads to limited pose estimation accuracy.
Therefore, we propose to predict per-view coarse geometry (in the form of a sparse point cloud, i.e., predicting one 3D point for each patch token) that is supervised to be consistent with the NeRF geometry, allowing us to obtain the camera poses with a PnP solver given the 3D-2D correspondences from the per-patch predicted points and patch centers.

In particular, from each image patch token output by the transformer $\ImgToken_{i,j}$, we use an MLP to predict a 3D point and the prediction confidence:
\begin{equation}
\ImgPoint_{i,j}, \Imgalpha_{i,j}, \ImgConf_{i,j}  = \text{MLP}_{\ImgToken}(\ImgToken_{i,j}),
\end{equation}
where $\ImgPoint_{i,j}$ represents the 3D point location on the object seen through the central pixel of the image patch, $\Imgalpha_{i,j}$ is the pixel opacity that indicates if the pixel covers the foreground object, and $\ImgConf_{i,j}$ is an additional confidence weight used to determine the point's contribution to the PnP solver.

Note that in training stage, where the ground-truth camera poses are known, the central pixel's point location and opacity can also be computed from a NeRF as done in previous work \citep{mildenhall2020nerf}. This allows us to enforce the consistency between the per-patch point estimates and the triplane NeRF geometry with following losses:
\begin{equation}
\label{eqn:pointloss}
\Loss_\ImgPoint = \sum\nolimits_{i,j}\|\ImgPoint_{i,j}-\bar{\Pos}_{i,j}\|^2, \quad \Loss_\Imgalpha = \sum\nolimits_{i,j}(\Imgalpha_{i,j}- (1 - \bar{\Trans}_{i,j}))^2,
\end{equation}
where $\bar{\Pos}$ and $\bar{\Trans}$ are computed along the pixel ray (marched from the ground-truth camera poses) using the volume rendering weights in Eqn.~\ref{eqn:nerfrendering} by
\begin{equation}
\bar{\Pos} = \sum_{k=1}^{K} \Trans_{k-1}(1-\exp (-\Dens_{k} \delta_{k}))\Pos_k, \quad \bar{\Trans} = \Trans_{K} = \exp(-\sum_{k'=1}^{K} \Dens_{k'} \delta_{k'}).
\end{equation}
Here $\bar{\Pos}$ represents the expected 3D location and $\bar{\Trans}$ is the final volume transmittance $\Trans_{K}$. Essentially, we distill the geometry of our learnt NeRF reconstruction to supervise our per-view coarse point cloud prediction in an online manner, as we only use multi-view posed images to train our model without accessing 3D ground-truth. This online distillation is critical to stabilize the differentiable PnP loss mentioned later in Eq.~\ref{eq:KL_est}, without which we find the training tend to diverge in our experiments.

When $\ImgPoint_{i,j}$ and $\Imgalpha_{i,j}$ are estimated, we can already compute each pose $\Pose_i$ with a standard weighted PnP solver that solves
\begin{align}\label{eq:pnp}
& \argmin_{\Pose_i = [R_i, t_i]} \frac{1}{2} \sum_{j=1}^M 
 \ProjLoss(\Pose_i, \ImgPoint_{i,j},  \beta_{i,j}), \\
& \ProjLoss(\Pose_i, \ImgPoint_{i,j}, \Imgalpha_{i,j}) =  \beta_{i, j} \| \mathcal{P}(R_i \cdot \ImgPoint_{i,j} + t_i ) - \ImgPixel_{i,j}\|^2, \\
& \beta_{i, j}=\Imgalpha_{i,j} \ImgConf_{i,j},
\end{align}

where $\ImgPixel_{i,j}$ is the 2D central pixel location of the patch, $[R_i,t_i]$ are the rotation and translation components of the pose $\Pose_i$, $\mathcal{P}$ is the projection function with camera intrinstics involved, and $\ProjLoss(\cdot)$ represents the pixel re-projection error weighted by predicted opacity and PnP confidence.
Here, the predicted opacity values are used to weigh the errors to prevent the non-informative white background points from affecting the pose prediction. 

However, computing the solution of PnP is a non-convex problem prone to local minimas. 
Therefore, we further apply a robust differentiable PnP loss, proposed by EPro-PnP~\citep{chen2022epro}~\footnote{We take the public implementation in \url{https://github.com/tjiiv-cprg/EPro-PnP}.}, to regularize our pose prediction, leading to much more accurate results (See Tab.~\ref{tab:ablation_pose}). This loss is expressed by
\begin{equation}\label{eq:KL_est}
\Loss_{\Pose_i} = \frac{1}{2} \sum\nolimits_{j}  \ProjLoss(\Pose_{i}^\text{gt}, \ImgPoint_{i,j}, \beta_{i,j})  + \log \int \exp \left( -\frac{1}{2} \sum\nolimits_{j}  \ProjLoss(\Pose_i,\ImgPoint_{i,j}, \beta_{i,j}) \right) \mathrm{d}\Pose_i,
\end{equation}

where the first term minimizes the reprojection errors of the predicted points with the ground-truth poses and the second term minimizes the reprojection errors with the predicted pose distribution using Monte Carlo integral; we refer readers to the EPro-PnP paper~\citep{chen2022epro} for details about computing the integral term. 
This differentiable PnP loss, combined with our point prediction losses (in Eqn.~\ref{eqn:pointloss}), leads to plausible per-patch point location and confidence estimates, allowing for accurate final pose prediction.

\subsection{Loss functions and implementation details}
\label{sec:loss_and_details}
{\bf Loss.} Combining all losses (Eqn.~\ref{eq:rendering_loss},\ref{eqn:pointloss},\ref{eq:KL_est}), our final training objective is 
\begin{equation} \label{eq:total_loss}
    \Loss = \Loss_\Color + \gamma_\ImgPoint\Loss_\ImgPoint + \gamma_\Imgalpha\Loss_\Imgalpha + \gamma_\Pose\sum\nolimits_{i=2}^M\Loss_{\Pose_i},
\end{equation}
where $\Loss_\Color$ represents the rendering loss and $\gamma_\ImgPoint$, $\gamma_\Imgalpha$, $\gamma_\Pose$ are the weights for individual loss terms related to per-view coarse geometry prediction, opacity prediction and differentiable PnP loss.

{\bf Implementation details.}
Our single-stream transformer model consists of 36 self-attention layers. We predict triplane of shape $H_T = W_T = 64, D_T=32$. In order to decrease the tokens used in transformer, the triplane tokens used in transformer is $3072=3\!\times\! 32 \! \times \! 32$ and will be upsampled to 64 with de-convolution, similar to LRM~\citep{hong2023lrm}. We set the loss weights $\gamma_C', \gamma_C''$ (Eq.~\ref{eq:rendering_loss}), $\gamma_\ImgPoint,\gamma_\Imgalpha,\gamma_\Pose$ (Eq.~\ref{eq:total_loss}) to $1, 2, 1, 1, 1$, respectively.  We use AdamW~\citep{loshchilov2017decoupled} ($\beta_1 = 0.9, \beta_2 = 0.95$) optimizer with weight decay $0.05$ for model optimization. The initial learning rate is zero, which is linearly warmed up to  $4\times 10^{-4}$
for the first 3k steps and then decay to zero by cosine scheduling. The batch size per GPU is 8. Training this model for 40 epochs takes 128 Nvidia A100 GPUs for about one week.
We use the deferred back-propagation technique~\citep{zhang2022arf} to save GPU memory in NeRF rendering.
For more implementation details, please refer to Sec.~\ref{sec:additional_details} of the appendix.

\section{Experiments}
\label{sec:experiments}

\subsection{Experimental settings} \label{sec:experiments_details}

{\bf Training datasets.} Our model only requires multi-view posed images to train. To construct a large-scale multi-view posed dataset, we use a mixture of multi-view posed renderings from Objaverse~\citep{objaverse} and posed real captures from MVImgNet~\citep{yu2023mvimgnet}. We render the Objavere dataset following the same protocol as LRM~\citep{hong2023lrm} and DMV3D~\citep{xu2023dmv3d}: each object is normalized to $[-1, 1]^3$ box and rendered at 32 random viewpoints. We also preprocess the MVImgNet captures to crop out objects, remove background~\footnote{Mask removal tool: \url{https://github.com/danielgatis/rembg}}, and normalizing object sizes in the same way as LRM and DMV3D. In total, we have multi-view images of $\sim$1 million objects in our training set: $\sim$730k from Objaverse, $\sim$220k from MVImgNet.

{\bf Evaluation datasets.} To evaluate our model's cross-dataset generalization capability, 
we utilize a couple of datasets, including OmniObject3D~\citep{wu2023omniobject3d}, Google Scanned Objects (GSO)~\citep{gso}, Amazon Berkeley Objects (ABO)~\citep{abo}, Common Objects 3D (CO3D)~\citep{reizenstein21co3d}, and DTU~\citep{aanaes2016dtu}. For OmniObject3D, GSO, ABO datasets, we randomly choose 500 objects for assessing our model's performance given sparse images as inputs. We render out 5 images from randomly selected viewpoints for each object; to ensure view sparsity, we make sure viewing angles between any two views are at least 45 degrees. We feed randomly-chosen 4 images to our model to predict a NeRF and poses, while using the remaining 1 to measure our novel-view rendering quality. For CO3D dataset, we use the 400 held-out captures provided by RelPose++~\citep{lin2023relposepp}, which covers 10 object categories. To remove background, we use the masks included in the CO3D dataset. However, we note that these masks can be very noisy sometimes, negatively affecting our model's performance and the baseline RelPose++ (mask variant). We randomly select 4 random input views for each capture.
For DTU dataset, we take the 15 objects with manually annotated masks provided by IDR~\citep{yariv2020multiview}; for each object, we randomly select 8 different combinations of four input views, resulting in a total of 120 different testing inputs.

{\bf Baselines.} As our \ourmethod{} can do joint pose and shape estimation, we evaluate its performance against baselines on both tasks. For the pose estimation task, we compare \ourmethod{} with FORGE~\citep{jiang2022forge}, RelPose++~\citep{lin2023relposepp}, and the SfM-based method HLoc~\citep{sarlin2019hloc, schonberger2016structure}. We also compare with FORGE in terms of the reconstruction quality. We did not compare with SparsePose~\citep{sinha2023sparsepose} as there is no public source code available. SRT~\citep{srt22} is geometry-free and does not directly predict shapes like us; hence we did not compare with it due to this clear distinction in problem scopes.

{\bf Metrics.} Since we only care about relative pose estimation in the pose estimation task, we use pair-wise relative pose errors as our metric: for each image pair in the input image set, we measure the rotation part of the relative pose by computing the error as the minimal rotation angle between the prediction and ground-truth. We also report the percentage of image pairs with relative rotation errors below thresholds $15^\circ$ and $30^\circ$. The translation part of the predicted relative pose is measured by its absolute difference 
from the ground-truth one. We evaluate the reconstruction quality by comparing renderings of our reconstructed NeRF using both \textit{predicted} input-view poses and \textit{ground-truth} novel-view poses against the ground-truth. We report the PSNR, SSIM and LPIPS~\citep{zhang2018perceptual} metrics for measuring the image quality.
We use 4 images as inputs for each object when comparing the performance of different methods.

\subsection{Experiment results}
\label{sec:experiment_results}
\subsubsection{Pose prediction quality}

\begin{table}[htb]
    \centering
    \caption[]{On pose prediction task, we compare cross-dataset generalization to OmniObject3D~\citep{wu2023omniobject3d}, GSO ~\citep{gso}, ABO~\citep{abo}, CO3D~\citep{reizenstein21co3d}, DTU~\citep{aanaes2016dtu} with baselines FORGE~\citep{jiang2022forge}, HLoc~\citep{sarlin2019hloc},  RelPose++~\citep{lin2023relposepp}. Note that RelPose++ is trained on CO3D training set; hence its numbers on CO3D test set are not exactly cross-dataset performance. On OmniObject3D, GSO, ABO where background is white in the rendered data, we evaluate the \textit{w/o bg} variant of RelPose++, while on CO3D and DTU where real captures contain background, we evalute its \textit{w/ bg} variant.
    }
    \begin{tabular}{c|c|c|c|c}
    \hline
    
    \multicolumn{5}{c}{\bf OmniObject3D} \\
    \hline
     Method & R. error $\downarrow$ & Acc.@$15^{\circ}$ $\uparrow$ & Acc.@$30^{\circ}$ $\uparrow$ & T. error $\downarrow$ \\
    \hline
    {\small FORGE} &  71.06 & 0.071 & 0.232 & 0.726 \\
    {\small HLoc (F. rate 99.6\%)} & 98.65 & 0.083 & 0.083 & 1.343  \\
    {\small RelPose++ (w/o bg)} & 69.22 & 0.070 & 0.273 & 0.712 \\
    {\small Ours} & {\bf 6.32} & {\bf 0.962} & {\bf 0.990} & {\bf 0.067} \\
    \hline
    
    \multicolumn{5}{c}{\bf GSO} \\
    \hline
    Method & R. error $\downarrow$ & Acc.@$15^{\circ}$ $\uparrow$ & Acc.@$30^{\circ}$ $\uparrow$ & T. error $\downarrow$ \\
    \hline
    {\small FORGE } & 103.81 & 0.012 & 0.056 & 1.100  \\
    {\small HLoc (F. rate 97.2\%)} & 97.12 & 0.036 & 0.131 & 1.199 \\
    {\small RelPose++ (w/o bg)} & 107.49 & 0.037 & 0.098 & 1.143 \\
    {\small Ours} & {\bf 3.99} & {\bf 0.956} & {\bf 0.976} & {\bf 0.041}\\
    \hline
    
    \multicolumn{5}{c}{\bf ABO} \\
    \hline
    Method & R. error $\downarrow$ & Acc.@$15^{\circ}$ $\uparrow$ & Acc.@$30^{\circ}$ $\uparrow$ & T. error $\downarrow$ \\
    \hline
    {\small FORGE } & 105.23 & 0.014 & 0.059 & 1.107  \\
    {\small HLoc (F. rate 98.8\%)} & 94.84 & 0.067 & 0.178 & 1.302 \\
    {\small RelPose++ (w/o bg)} & 102.30 & 0.060 & 0.144 & 1.103 \\
    {\small Ours} & {\bf 16.27} & {\bf 0.865} & {\bf 0.885} & \textbf{0.150} \\
    \hline

    \multicolumn{5}{c}{\bf CO3D} \\
    \hline
    Method & R. error $\downarrow$ & Acc.@$15^{\circ}$ $\uparrow$ & Acc.@$30^{\circ}$ $\uparrow$ & T. error $\downarrow$ \\
    \hline
    {\small FORGE} & 77.74 & 0.139 & 0.278 & 1.181  \\
    {\small HLoc (F. rate 89.0\%)} & 55.87 & 0.288 & 0.447 & 1.109  \\
    {\small RelPose++ (w/ bg)} & 28.24 & 0.748 & 0.840 & 0.448 \\
    {\small Ours} & \textbf{15.53} & \textbf{0.850} & \textbf{0.899} & \textbf{0.242}  \\
    \hline

    \multicolumn{5}{c}{\bf DTU} \\
    \hline
    Method & R. error $\downarrow$ & Acc.@$15^{\circ}$ $\uparrow$ & Acc.@$30^{\circ}$ $\uparrow$ & T. error $\downarrow$\\
    \hline
    {\small FORGE} & 78.88 & 0.046 & 0.188 & 1.397\\
    {\small HLoc (F. rate 47.5\%)} & 11.84 & 0.725 & 0.915 & 0.520\\
    {\small RelPose++ (w/ bg)} & 41.84 & 0.369 & 0.657 & 0.754\\
    {\small Ours} & \textbf{10.42} & \textbf{0.900} & \textbf{0.951} & \textbf{0.187} \\
    \hline

    \end{tabular}
    \label{tab:pose_compare_large}
\end{table}

As shown in Tab.~\ref{tab:pose_compare_large}, our model achieves state-of-the-art results in pose estimation accuracy and rendering quality given highly sparse input images on unseen datasets including OmniObjects3D, ABO, GSO, CO3D, and DTU, consistently outperforming baselines by a large margin across all datasets and all metrics. This is an especially challenging evaluation setting, as we are assessing the cross-dataset generalization capability of different methods, which reflects their performance when deployed in real-world applications. In this regard, we directly use the pretrained checkpoints from baselines, including~FORGE~\citep{jiang2022forge}, HLoc~\citep{sarlin2019hloc} and RelPose++~\citep{lin2023relposepp}, for comparisons.

On OmniObject3D, GSO, ABO datasets where the input images are explicitly sparsified (see Sec.~\ref{sec:experiments_details}), we achieve an average of 14.6x reduction in rotation error for the predicted poses compared with FORGE, while the average rotation error reductions are 15.3x compared with HLoc and 14.7x compared with RelPose++. As FORGE expects input images to have black background, we replace the white background in our rendered images  with a black one using the rendered alpha mask before feeding them into FORGE. RelPose++, however, has two variants: one trained on images with background (\textit{w/ bg}) and the other trained on images without backgrund (\textit{w/o bg}). We evaluate the \textit{w/o bg} variant on these datasets featuring non-informative white background. In addition, we observe that HLoc has a very high failure rate (more than 97\%) on these very sparse inputs, due to that matching features is too hard in this case; this also highlights the difficulty of pose prediction under extremely sparse views, and the contributions this work to the area. 

On the held-out CO3D test set provided by RelPose++, our rotation error is 5x smaller than FORGE, 3.6x smaller than HLoc, and 1.8x smaller than RelPose++ (\textit{w/ bg}). Note that FORGE, HLoc and our method are all tested on input images with background removed. The inaccurate foreground masks provided by CO3D can negatively influence these methods' performance; this can explain our performance degradation from datasets like OmniObject3D to datasets like CO3D. It will be interesting to explore ways to extend our method to handle background directly in future work. We also note that CO3D captures may not cover the objects in 360, hence we do not sparsify the input poses but instead use randomly selected views; evaluation on this dataset may not reflect different models' performance on highly sparse inputs. In addition, RelPose++ is trained on the CO3D training set while the other methods, including ours, are not. On DTU dataset where none of the methods are trained on, we achieve 4x rotation error reduction than RelPose++ (\textit{w/ bg}), and 7.6x reduction than FORGE, showing much better generalization capability than these methods. Interestingly, as we do not sparsify input views intentionally on this data for the same reason as CO3D, HLoc, which relies on traditional SfM~\citep{schonberger2016structure} to solve poses, can produce reasonably accurate pose estimations if enough features are correctly matched (failure rate is 47.5\%); however, it's performance is still far worse than ours, especially on the metric $Acc. @15^\circ$ that measures the percentage of pair-wise rotation errors below the threshold of 15 degrees.

We attribute our model's success to the prediction of
both camera poses and object shapes at the same time, where the synergy of the two tasks are exploited by the self-attention mechanism. The generic shape prior learned on the large Objaverse and MVImgNet datasets differentiate our method from other methods, as we find it particularly helpful in estimating camera parameters given sparse inputs (See Sec.~\ref{sec:ablation}).
Prior methods like RelPose++ failed to utilize this synergy, as they solve for poses directly from images without simultaneously reconstructing the 3D object.
FORGE designed a learning framework to introduce shape prior into the pose estimation process, but its training process is composed of six stages, which seems fragile and not easy to scale up, compared with our single-stream transformer design.  Therefore it shows much weaker cross-dataset generalization capability than our method.  This said, we also acknowledge that if one can successfully scale up the training of the baseline RelPose++ and FORGE on large-scale datasets, their performance can also be improved compared with their pretrained model using limited data. We show one such experiment where we re-train RelPose++ on Objaverse data in appendix~\ref{sec:scale_up_relposepp}; however, our model, trained on exactly the same Objaverse data, still outperforms this re-trained baseline by a wide margin in terms of pose prediction accuracy on various evaluation datasets, demonstrating the superiority of our method. We leave the investigation of scaling up FORGE to future work due to its complex pipeline.

Although the SfM method HLoc solves for poses and 3D shape (in the form of sparse point cloud) at the same time, it relies on feature matching across views which is extremely challenging in the case of sparse-views; hence it performs poorly in our application scenario.

\subsubsection{Reconstruction quality}

\begin{figure} 
\includegraphics[width=0.98\textwidth]{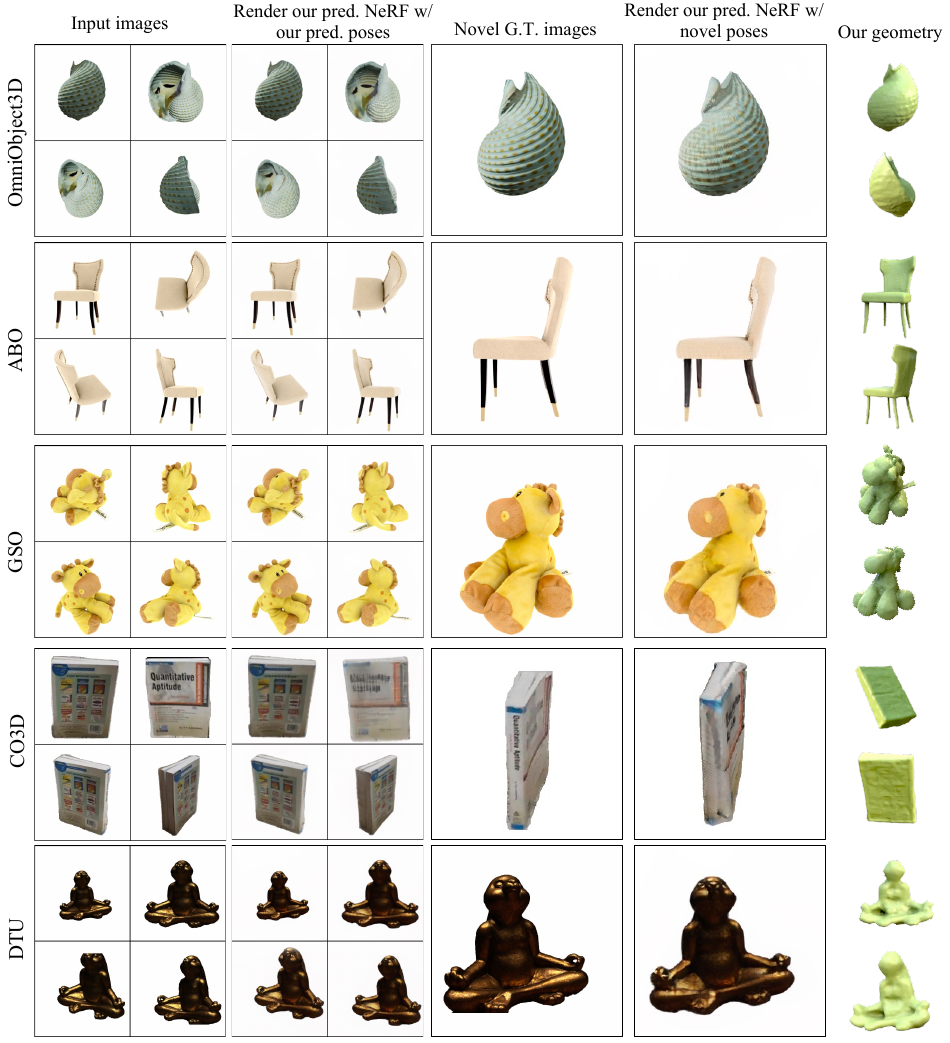}
\caption{Cross-dataset generalization to unseen OmniObject3D~\citep{wu2023omniobject3d}, GSO~\citep{gso} and ABO~\citep{abo} datasets. Renderings 
of our \textit{predicted} NeRF at \textit{predicted} poses (second column) closely match the input unposed images (first column), demonstrating the excellent accuracy of both predictions; we also show novel-view rendering of our reconstructed NeRF (fourth column) and the corresponding ground-truth (third column) to show our high-quality NeRF reconstruction, from which we can also easily extract meshes (last column) by fusing the mult-view RGBD images rendered from NeRF using RGBD fusion~\citep{rgbd-fusion}. More visual examples can be found in Fig.~\ref{fig:addional_qualitative_results} in the appendix. 
}
\label{fig:our_main_results}
\end{figure}

We use the surrogate view synthesis quality to compare the quality of our reconstructed NeRF to that of FORGE~\citep{jiang2022forge}. To isolate the influence of inaccurate masks on measuring the view synthesis quality, we evaluate on unseen OmniObject3D, GSO, and ABO datasets and compare with the baselines FORGE.
In this experiment, we use the same input settings as in the above pose prediction comparisons. We use PSNR, SSIM~\citep{wang2004image}, and LPIPS~\citep{zhang2018perceptual} as image metrics.

As shown in Tab.~\ref{tab:novel_view_synthesis}, our \ourmethod{} achieves an average PSNR of 24.8 on OmniObject3D, GSO, and ABO datasets, while the baseline FORGE's average PSNR is only 13.4.
This shows that our model generalizes very well and produce high-quality reconstructions on unseen datasets while FORGE does not. Note that we actually feed images with black background into FORGE, and evaluate PSNR using images with black background; this is, in fact, an evaluation setup that bias towards FORGE, as images with black background tends to have higher PSNR than those with white background.

On the other hand, we think there's an important objective to fulfill in the task of joint pose and NeRF prediction; that is, the predicted NeRF, when rendered at predicted poses, should match well the input unposed images. This is an objective complimentary to the novel view quality and requiring  accurate predictions of both poses and NeRF. We show in Tab.~\ref{tab:novel_view_synthesis} that FORGE does poorly on this goal evidenced by the low PSNR scores, especially on the GSO and ABO datasets. In contrast, we perform much better.

In general, our model learns a generic shape prior effectively from massive multi-view datasets including Objaverse and MVImgNet, thanks to its scalable single-stream transformer design. FORGE's multi-stage training, though, is challenging to scale due to error accumulation across stages. Fig.~\ref{fig:our_main_results}  qualitatively shows the high-quality NeRF reconstruction and accurate pose prediction from our model. Renderings of the our predicted NeRF using our predicted poses closely match the input images, and the novel view rendering resembles the ground-truth a lot. We also demonstrate high-quality extracted meshes from our reconstructed NeRF; the meshes are extracted by first rendering out 100 RGBD images uniformly distributed in a sphere and then fusing them using RGBD fusion~\citep{rgbd-fusion}.

\begin{table}[htb]
    \caption[]{On 3D reconstruction task, we compare novel view synthesis quality with baseline FORGE~\citep{jiang2022forge} on OmniObject3D~\citep{wu2023omniobject3d}, GSO ~\citep{gso}, ABO~\citep{abo} datasets. Neither methods are trained on these evaluation datasets. Note that both methods predict input cameras' poses and hence the predicted NeRF are aligned with their own predicted cameras; we align the predicted input cameras to the ground-truth one, and transform the reconstructed NeRF accordingly before rendering them with novel-view cameras for computing the image metrics. We show evaluate renderings of the predicted NeRF at the predicted camera poses against the inputs to show how consistent both predictions are in terms of matching inputs. 
    }
\resizebox{\linewidth}{!}{
    \centering
    \begin{tabular}{c|ccc|ccc|ccc}
    \hline
    &\multicolumn{3}{c|}{OmniObject3D} & \multicolumn{3}{c|}{Google Scanned Objects} & \multicolumn{3}{c}{Amazon Berkeley Objects} \\
    \hline
    Method & PSNR $\uparrow$ & SSIM $\uparrow$ & LPIPS $\downarrow$ & PSNR $\uparrow$ & SSIM $\uparrow$ & LPIPS $\downarrow$ & PSNR $\uparrow$ & SSIM $\uparrow$ & LPIPS $\downarrow$ \\
    \hline
     & \multicolumn{9}{c}{Evaluate renderings of our predicted NeRF at novel-view poses} \\
    \hline
     {FORGE} & 17.95 & 0.800 & 0.215 & 11.43 & 0.754 & 0.760 & 10.92 & 0.669 & 0.325 \\
     {Ours} & \textbf{23.02} & \textbf{0.877} & \textbf{0.083} & \textbf{25.04} & \textbf{0.879} & \textbf{0.096} & \textbf{26.23} & \textbf{0.887} & \textbf{0.097} \\
     \hline
    & \multicolumn{9}{c}{Evaluate renderings of our predicted NeRF at our predicted poses} \\
     \hline
      {FORGE} & 19.03 & 0.829 & 0.189 & 11.90 & 0.760 & 0.202 & 11.32 & 11.32 & 0.209 \\
     {Ours} & \textbf{27.27} & \textbf{0.916} & \textbf{0.054} & \textbf{27.01} & \textbf{0.914} & \textbf{0.0645} & \textbf{27.19} & \textbf{0.894} & \textbf{0.083}  \\
     \hline
    \end{tabular}
}
    \label{tab:novel_view_synthesis}
\end{table}

\begin{figure}[htb] 
\centering
\includegraphics[width=0.85\textwidth]{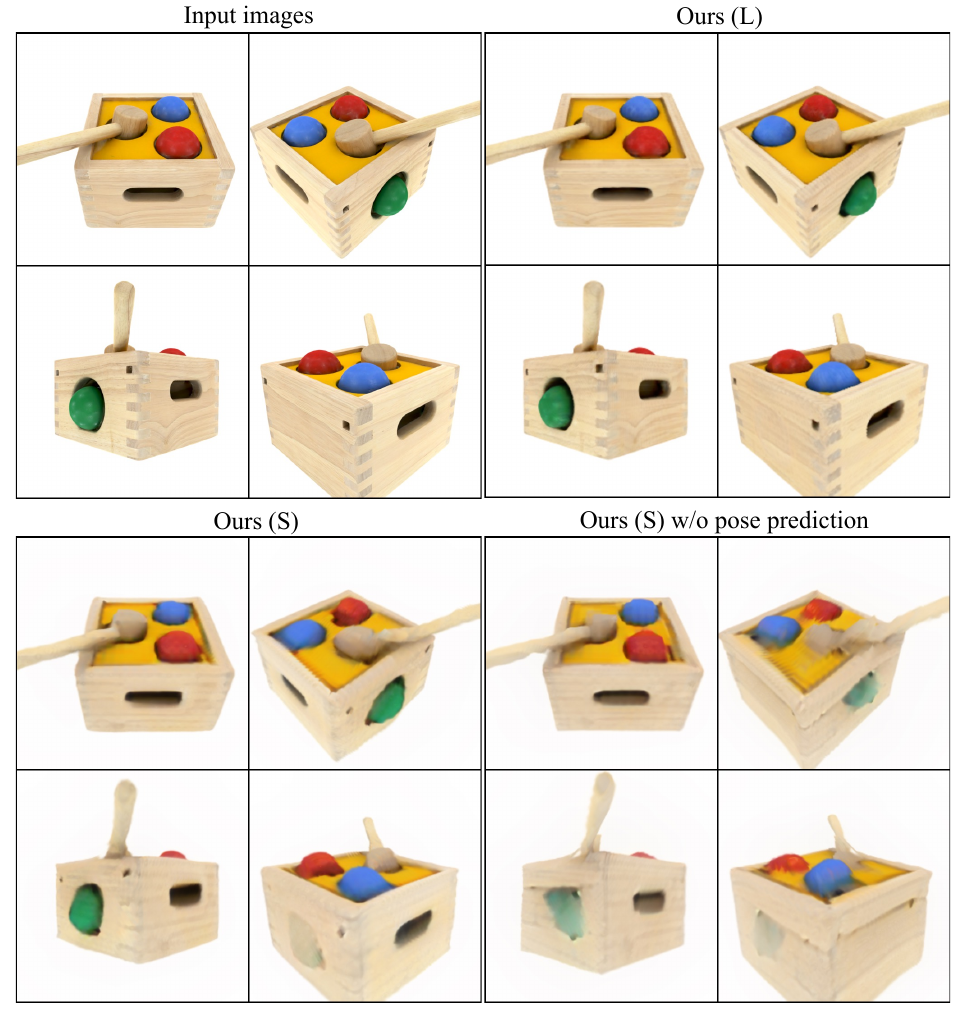}
\caption{Ablation studies on GSO data~\citep{gso}. `Ours (L)' results in highest reconstruction quality with sharpest details, while reducing the model size (`Ours (S)') causes the texture to become blur. Further removing pose prediction branch (`Ours (S) w/o pose prediction') makes the texture even worse. Note that for a fair comparison of different ablation variants, especially the one without pose prediction, we render out our reconstructed NeRF using the same \textit{ground-truth} poses corresponding to input images (as opposed to predicted ones).  %
}
\label{fig:ablation_main}
\end{figure}

\subsection{Robustness Tests}
{\bf Variable number of input views.}  Our model naturally supports variable number of input views as a result of the transformer-based architecture. We test our model's performance on variable number of input images on the 500 selected GSO objects (see Sec.~\ref{sec:experiments_details}). As shown in Tab.~\ref{tab:ablation_varing_views}, with decreased number of views, we observe a consistent drop in reconstruction quality and pose prediction quality, but the performance degradation is acceptable. Note that for pose evaluation, we only evaluate
the relative pose errors of the first two views for fair comparison. PSNR$_{input}$ reflects how well our model's predicted NeRF and poses can explain the input images, while the PSNR$_{all}$ is an aggregated metrics including both input views and held-out novel views (we have 4 views in total for each object).

\begin{table}[htb]
    \caption[]{Inference on variable number of input views on unseen GSO dataset using our \ourmethod{} trained on 4 views (no re-training or fine-tuning is involved). For pose evaluation, we only evaluate the relative pose erros of the first two views for fair comparison.}
    \centering
    \begin{tabular}{c|c|c|c|c|c}
    \hline
    \#Views & R. error & Acc.@$15^{\circ}$ & Acc.@$30^{\circ}$ & PSNR$_{\rm input}$ & PSNR$_{\rm all}$\\
    \hline
   {\small 4} & \textbf{4.19} & \textbf{0.956} & \textbf{0.974} & 27.76 & \textbf{27.76} \\
   {\small 3} & 5.83 & 0.946 & 0.962 & 27.59 & 26.76 \\
   {\small 2} & 10.38 &	0.886 & 0.924 & 27.35 & 24.87 \\
   {\small 1} & - & - & - & \textbf{29.27} & 21.56 \\
    \hline
    \end{tabular}
    \label{tab:ablation_varing_views}
\end{table}

{\bf Imperfect segmentation masks.} In this experiment we add noises on the input segmentation masks by adding different levels of elastic transform~\citep{simard2003best}.
As shown in Tab.~\ref{tab:ablation_inperfect_masks}, we can see that our model is robust to certain level of noise, but its performance drop significantly when the masks are very noisy. This is also aligned with the observation that the inaccurate masks provided by CO3D~\citep{reizenstein21co3d} can harm our model's performance on it, e.g., the Couch category in Tab.~\ref{tab:pose_compare_co3d} of the appendix. Note that PSNR$_{g.t.}$ reflects how well renderings of our predicted NeRF using ground-truth input poses match the input images, while  PSNR$_{pred}$ measures how well renderings of our predicted NeRF using our pose predictions match the inputs.

\begin{table}[htb]
    \caption[]{Inference on images with varying level of segmentation mask errors on unseen GSO dataset using our \ourmethod.%
    }
    \centering
    \begin{tabular}{c|c|c|c|c|c|c}
    \hline
    Noise level & R. error & Acc.@$15^{\circ}$ & Acc.@$30^{\circ}$ & T. error & PSNR$_{\rm g.t.}$ & PSNR$_{\rm pred.}$\\
    \hline
    {\small 0 } & \textbf{2.46} & \textbf{0.976} & \textbf{0.985} & \textbf{0.026} & \textbf{29.42} & \textbf{28.38} \\
    {\small 1} & 4.84 & 0.951 & 0.968 &	0.050 &	27.19 &	26.84 \\
    {\small 2} & 7.15 & 0.921 & 0.946 &	0.075 &	26.25 &	26.26 \\
    {\small 3} & 10.34 & 0.881 & 0.916 & 0.106 & 25.515 & 25.567 \\
    {\small 4} & 14.13 & 0.844 & 0.894 & 0.141 & 24.934 & 24.975 \\
    
    \hline
    
    \end{tabular}
    \label{tab:ablation_inperfect_masks}
\end{table}

\subsection{Ablation studies} \label{sec:ablation}

\begin{table}[htb]
    \centering
    \caption[]{Ablation study of model size and training objectives on the GSO dataset.}
    \begin{tabular}{c|c|c|c|c|c|c}
    \hline
    Setting & R. error & Acc.@$15^{\circ}$ & Acc.@$30^{\circ}$ & T. error & PSNR$_{\rm g.t.}$ & PSNR$_{\rm pred.}$\\
    \hline
    {\small Ours (L)} & \textbf{2.46} & \textbf{0.976} & \textbf{0.985} & \textbf{0.026} & \textbf{29.42} & \textbf{28.38} \\
    {\small Ours (S)} & 13.08 & 0.848 & 0.916 & 0.135 & 23.80 & 22.82 \\
    {\small - NeRF Pred. (S)} & 111.89 & 0.000 & 0.000 & 1.630 & - & - \\
    {\small - pose Pred. (S)} & - & - & - & - & 22.48 & - \\
    \hline
    \end{tabular}
    \label{tab:ablation_main}
\end{table}

In the ablation studies, we train our models with different settings on the synthetic Objaverse dataset~\citep{objaverse} and evaluate on GSO dataset~\citep{gso} to isolate the influence of noisy background removals. For better energy efficiency, we conduct ablations mostly on a smaller version of our model, dubbed as Ours (S). It has 24 self-attention layers with 1024 token dimension, and is trained on 8 A100 GPUs for 20 epochs ($\sim$100k iterations), which takes around 5 days. In addition, to show the scaling law with respect to model sizes, we train a large model (Ours (L)) on 128 GPUs for 100 epochs ($\sim$70k iterations).

{\bf Using smaller model.} `Ours (L)' outperforms the smaller one `Ours (S)' by a great margin in terms of pose prediction accuracy and NeRF reconstruction quality, as shown in Tab.~\ref{tab:ablation_main} and Fig.~\ref{fig:ablation_main}. It aligns with the recent findings that larger model can learn better from data~\citep{hong2023lrm}.

{\bf Removing NeRF prediction.} We evaluated two different settings without NeRF prediction: 1) using differentiable PnP for pose prediction as described in Sec.~\ref{sec:pose};  2) using MLP to directly predict poses from the concatenated patch features. For 1), we notice that the training becomes very unstable and tends to diverge in this case, as we find that our point loss (Eqn.~\ref{eqn:pointloss}; relying on NeRF prediction for supervision) helps stabilize the differentiable PnP loss (Eqn.~\ref{eq:KL_est}). For 2), we find that the predicted pose is almost random, as shown in Tab.~\ref{tab:ablation_main}; this indicates that the training and evaluation cases of highly sparse views (e.g., four images looking at the front, back, left- and right-side parts of an object) seem to pose a convergence challenge for a purely images-to-poses regressor when trained on the massive Objaverse dataset~\citep{objaverse}.

{\bf Removing pose prediction.} We find that jointly predicting pose helps the model learn better 3D reconstruction with sharper textures, as shown in Tab.~\ref{tab:ablation_main} (comparing `-pose Pred. (S)' and `Ours (S)') and Fig.~\ref{fig:ablation_main} (comparing `Our (S) w/o pose prediction' and `Ours (S)'). This could be that by forcing the model to figure out the correct spatial relationship of input views, we reduce the uncertainty and difficulty of shape reconstruction.

\subsection{Application}

{\bf Text/image-to-3D generation.} Since our model can reconstruct NeRF from 2-4 unposed images, it can be readily used in downstream text-to-3D applications to build highly efficient two-stage 3D generation pipelines. In the first stage, one can use geometry-free multi-view image generators, e.g., MVDream~\citep{shi2023MVDream}, Instant3D~\citep{li2023instant3d}, to generate a few images from a user-provided text prompt. Then the unposed generated images can be instantly lifted into 3D  by our \ourmethod{} with a single feed-forward inference (see Fig.~\ref{fig:teaser}). Or alternatively, one can generate a single image from text prompts using Stable Diffusion~\citep{rombach2022high}, feed the single image to image-conditioned generators, e.g., Zero-1-to-3~\citep{liu2023zero1to3}, Zero123++~\citep{shi2023zero123plus}, to generate at least one additional view, then reconstruct a NeRF from the multiple unposed images using our \ourmethod{}. In the latter approach, we can have a feed-forward single-image-to-3D pipeline as well, if the text-to-image step is skipped, as shown in Fig.~\ref{fig:teaser}.

\section{Conclusion}
\label{sec:conclusion}
In this work, we propose a large reconstruction model based on the transformer architecture to jointly estimate camera parameters and reconstruct 3D shapes in the form of NeRF. Our model employs self-attention to allow triplane tokens and image patch tokens to communicate information with each other, leading to improved NeRF reconstruction quality and robust per-patch surface point prediction for solving poses using a differentiable PnP solver. Trained on multi-view posed renderings of the large-scale Objaverse and real MVImgNet datasets, %
our model outperforms baseline methods by a large margin in terms of pose prediction accuracy and reconstruction quality. We also show that our model can be leveraged in downstream applications like text/image-to-3D generation.

{\bf Limitations.} Despite the impressive reconstruction and pose prediction performance of our model, there are a few limitations to be addressed in future works: 
1) First, we ignore the background information that might contain rich cues about camera poses, e.g., vanishing points, casting shadows, etc, while predicting camera poses. It will be interesting to extend our work to handle background with spatial warpings as in~\citep{zhang2020nerf++,barron2022mip}. 2) Second, we are also not able to model view-dependent effects due to our modelling choice of per-point colors, compared with NeRF~\citep{mildenhall2020nerf,verbin2022ref}. Future work will include recovering view-dependent appearance from sparse views. 3) The resolution of our predicted triplane NeRF can also be further increased by exploring techniques like coarse-to-fine modelling or other high-capacity compact representations, e.g., multi-resolution hashgrid~\citep{muller2022instant}, to enable more detailed geometry and texture reconstructions.  4) Our model currently assumes known intrinsics (see Sec.~\ref{sec:transformer}) from the camera sensor metadata or a reasonable user guess; future work can explore techniques to predict camera intrinscis as well.  5) Although our model is pose-free during test time, it still requires ground-truth pose supervision to train; an intriguing direction is to lift the camera pose requirement during training in order to consume massive in-the-wild video training data.

\paragraph{Ethics Statement.} 
The model proposed in this paper is a reconstruction model that can convert multi-view images to the 3D shapes. 
This techniques can be used to reconstruct images with human.
However, the current shape resolution is still relatively low which would not get accurate reconstruction of the face region/hand region.
The model is trained to be a deterministic model thus it is hard to leak the data used in training.
The users can use this model to reconstruct the shape of the images where there might be a commercial copyright of the shape.
This model also utilizes a training compute that is significantly larger than previous 3D reconstruction models.
Thus the model can potentially lead to a trend of pursuing large reconstruction models in the 3D domain, 
which further can introduce the environmental concerns like the current trend of large language model.

\paragraph{Reproducibility Statement.} 
We have elucidate our model design in the paper including the training architecture (transformer in Sec.~\ref{sec:transformer}, NeRF rendering in Sec.~\ref{sec:rendering}) the losses (pose loss in Sec.~\ref{sec:pose} and final loss in Sec.~\ref{sec:loss_and_details}).
The training details are shown in Sec.~\ref{sec:loss_and_details} and further extended in Appendix. 
We also pointed to the exact implementation of the Diff. PnP method in Sec.~\ref{sec:pose} to resolve uncertainty over the detailed implementation.
Lastly, we will involve in the discussion regarding implementation details of our paper.

\paragraph{Acknowledgement.}
We want to thank Nathan Carr, Duygu Ceylan, Paul Guerrero, Chun-Hao Huang, and Niloy Mitra for discussions on this project. We thank Yuan Liu for helpful discussions on pose estimation.

\bibliography{reference}
\bibliographystyle{reference}

\newpage

\appendix
\section{Appendix}

\begin{figure} 
\includegraphics[width=0.98\textwidth]{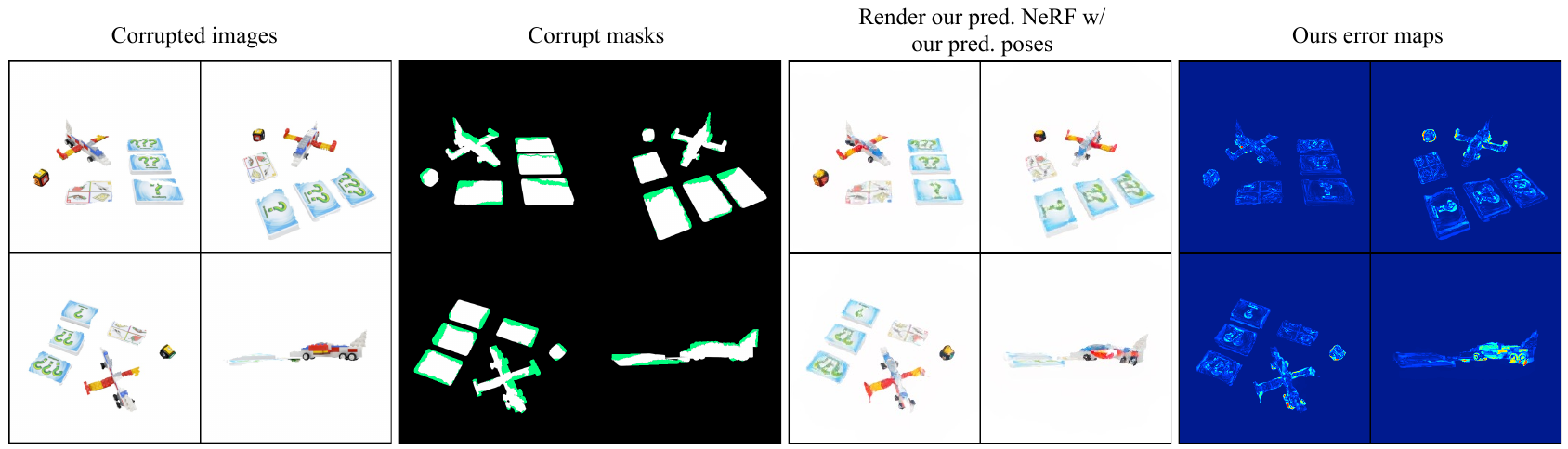}
\caption{Our \ourmethod{} is robust to small mask segmentation errors.}
\label{fig:robust_mask}
\end{figure}

\subsection{Visual comparisons of predicted camera poses}
\begin{figure} 
\includegraphics[width=0.98\textwidth]{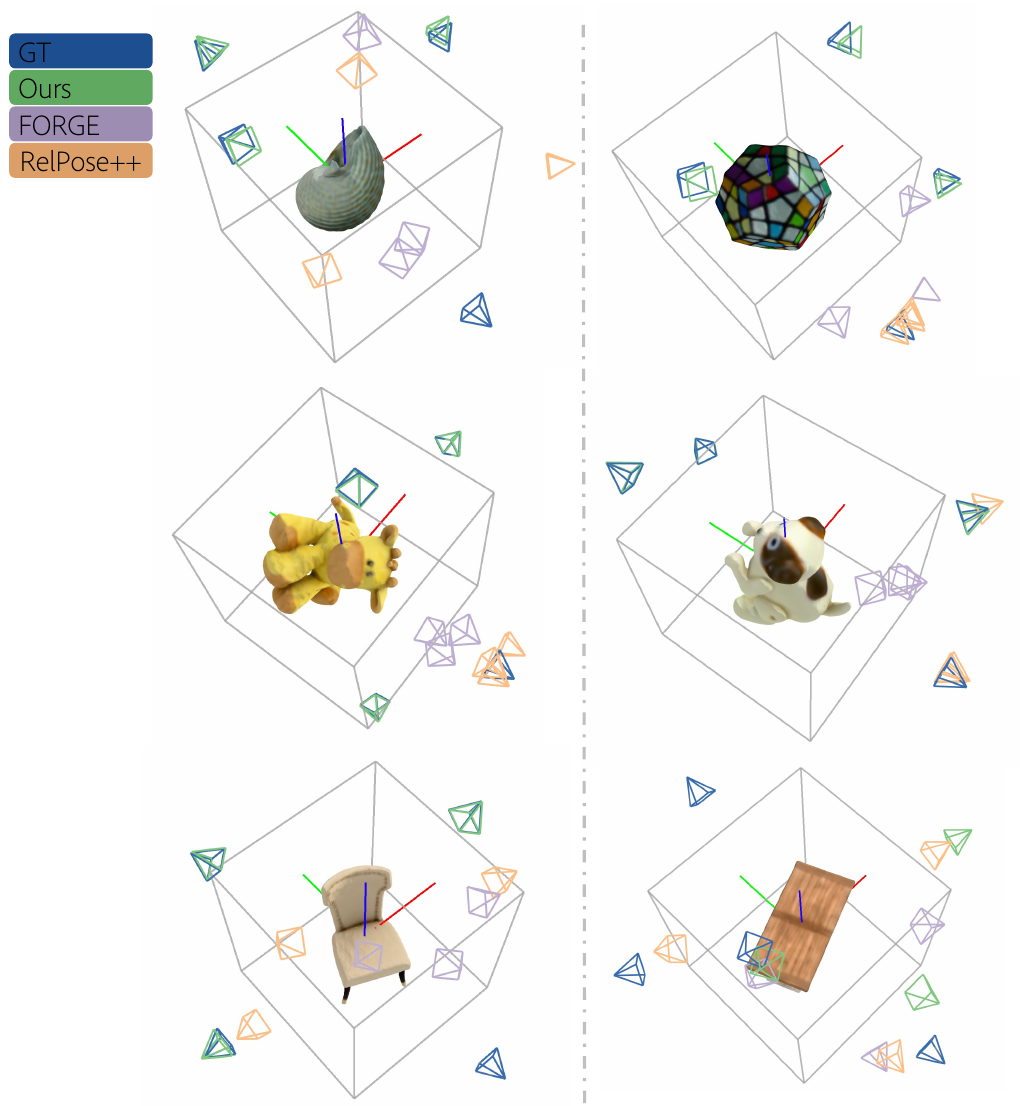}
\caption{Predicted poses from our method align much more closely with the ground-truth than those from baseline methods including FORGE~\citep{jiang2022forge}, RelPose++~\citep{lin2023relposepp}.}
\label{fig:pose_vis}
\end{figure}

In Fig.~\ref{fig:pose_vis}, we present visual comparisons of the predicted camera poses with our method and  
baseline methods. We can see that it's common for baseline methods FORGE~\citep{jiang2022forge} and RelPose++~\citep{lin2023relposepp} to make predictions significantly deviating from the ground truth and in some situations, their predicted camera poses can be even 
on the opposite side.  In contrast, our predicted poses closely align with 
the ground truth consistently.

\begin{figure} 
\includegraphics[width=0.98\textwidth]{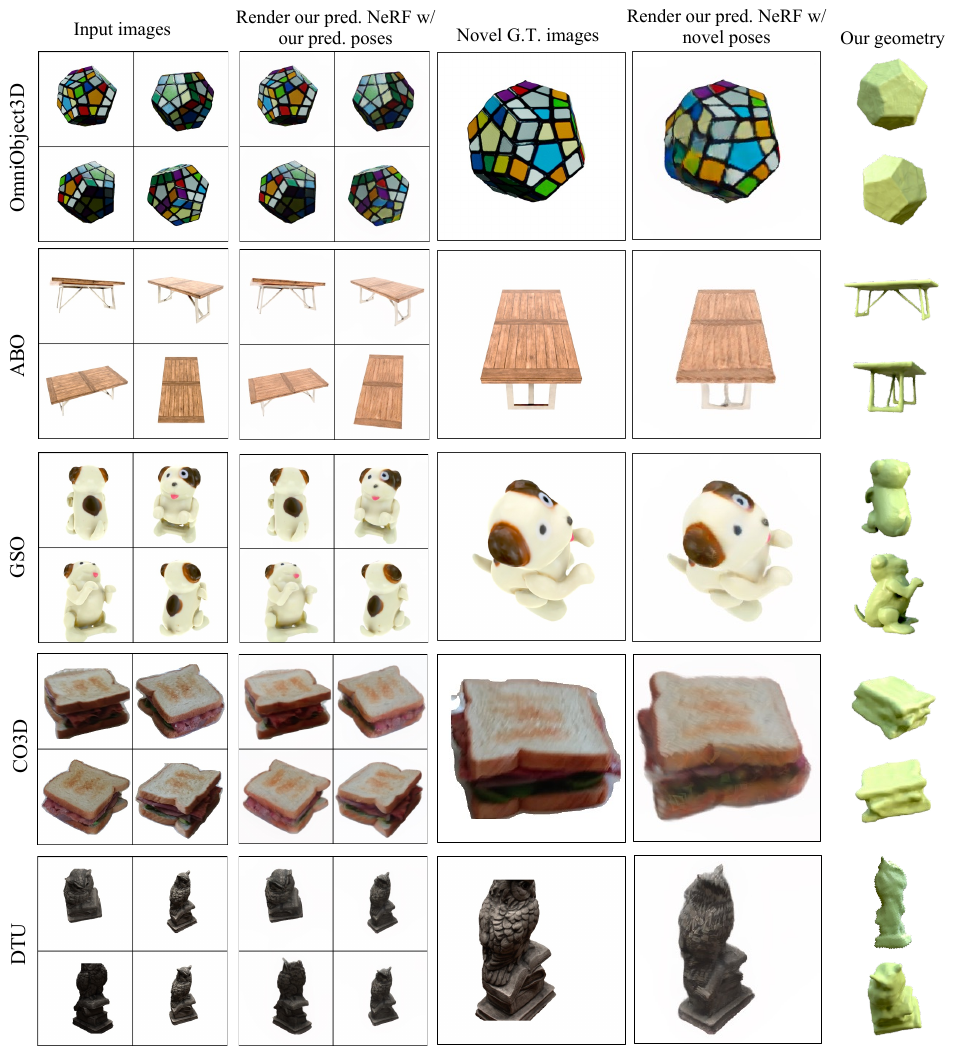}
\caption{Additional qualitative results of our model's cross-dataset generalization to unseen OmniObject3D~\citep{wu2023omniobject3d}, GSO~\citep{gso}, ABO~\citep{abo}, CO3D~\citep{reizenstein21co3d}, and DTU~\citep{aanaes2016dtu} datasets. }
\label{fig:addional_qualitative_results}
\end{figure}

\begin{table}[htb]
    \caption[]{Category-level comparison of pose prediction results with baseline RelPose++~\citep{lin2023relposepp} on CO3D dataset ~\citep{reizenstein21co3d}. We report the mean pose errors and (top two rows) and rotation accuracy@$15^{\circ}$ (bottom two rows) on 10 different test categories.}
\resizebox{\linewidth}{!}{
    \centering
    \begin{tabular}{c|c|c|c|c|c|c|c|c|c|c}
    \hline
         & Ball & Book & Couch & Fris. & Hot. & Kite & Rem. & Sand. & Skate. & Suit. \\
    \hline
    {\small RelPose++ (w/ bg)} & 30.29 & 31.34 & \textbf{24.82} & 34.01 & \textbf{21.61} & 50.18 & 32.00 & 30.84 & 36.91 & 14.13 \\
    {\small Ours} & \textbf{17.17} & \textbf{8.36} & 29.04 & \textbf{20.16} & 27.88 & \textbf{16.18} & \textbf{6.05} &	\textbf{15.92} &	\textbf{25.39} &	\textbf{12.03} \\
    \hline
    {\small RelPose++ (w/ bg)} & 0.613	& 0.782 & \textbf{0.787} & 0.742 & \textbf{0.742} & 0.570 & 0.767 & 0.697 & 0.630 & 0.893 \\
    {\small Ours} & \textbf{0.787} & \textbf{0.947} & 0.688 & \textbf{0.780} & 0.697 & \textbf{0.807} & \textbf{0.944} & \textbf{0.890} & \textbf{0.778} &	\textbf{0.923} \\
    \hline    
    \end{tabular}
}
    \label{tab:pose_compare_co3d}
\end{table}

\subsection{Additional Experiments}

\begin{figure}[ht]
\includegraphics[width=0.98\textwidth]{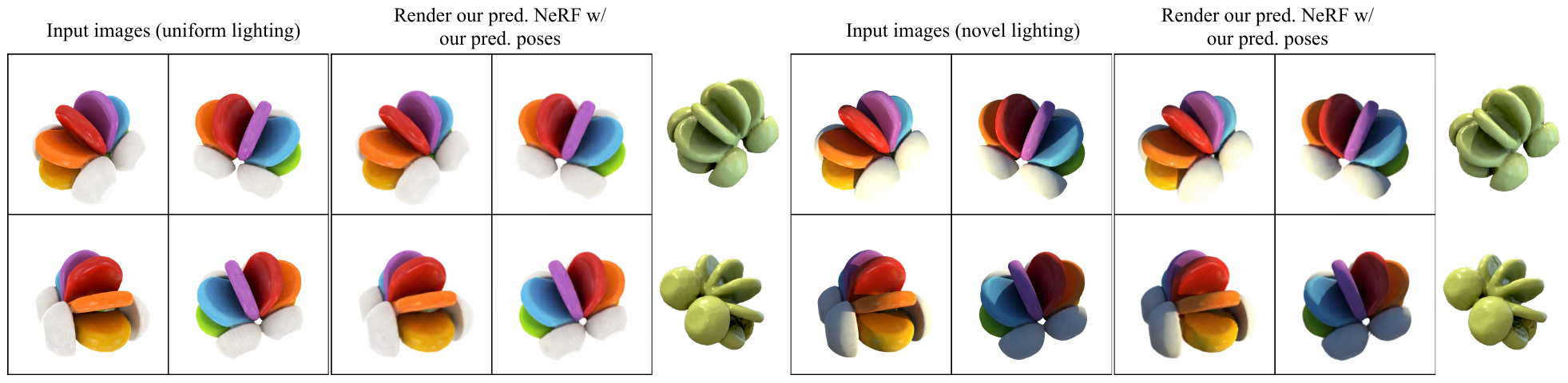}
\caption{Our \ourmethod{} is robust to illumination changes.}
\label{fig:robust_light}
\end{figure}

{\bf Robustness to novel environment lights.} 
We evaluate our model's robustness to different environment lights in Table~\ref{tab:ablation_novel_lighting}.
The evaluations are conducted in 100 object samples from GSO dataset~\citep{gso}.
Our model shows consistent results under different lighting conditions.
We also qualitatively shows our model robustness to different illuminations in Fig.~\ref{fig:robust_light}.  Note that PSNR$_{g.t.}$ reflects how well renderings of our predicted NeRF using ground-truth input poses match the input images, while  PSNR$_{pred}$ measures how well renderings of our predicted NeRF using our poses predictions match the inputs.

\begin{table}[htb]
    \caption[]{Evaluation results on GSO data with different novel environment lights. The evaluations are conducted in 100 objects samples. Note our synthesized multi-view training images are rendered using uniform light. Our method can generalize well to novel environment lights.}
    \centering
    \begin{tabular}{c|c|c|c|c|c|c}
    \hline
    Method & R. error & Acc.@$15^{\circ}$ & Acc.@$30^{\circ}$ & T. error & PSNR$_{\rm g.t.}$ & PSNR$_{\rm pred.}$\\
    \hline
    {\small Sunset} & 2.40 & 0.968 & 0.983 & 0.027 & 27.56 & 26.74 \\ 
    {\small Sunrise} & 2.22 &	0.985 &	0.993 &	0.024 &	27.17 & 26.21 \\
    {\small Studio} & 2.82 & 0.983 & 0.992 & 0.029 & 27.31 & 26.69 \\
    {\small Uniform} & 3.94 & 0.968 & 0.972 & 0.040 & 27.50 & 26.80 \\
    \hline
    \end{tabular}
    \label{tab:ablation_novel_lighting}
\end{table}

{\bf Ablations of pose prediction methods.}
We illustrate the effectiveness of our differentiable PnP pose prediction method in Tab.~\ref{tab:ablation_pose} by replacing it with alternative solutions. 
The first line `diff. PnP' is our model with small config, i.e., `Ours (S)'.
For other lines, we replace the `diff. PnP' with other alternatives.
`MLP pose (CLS token)' takes the [CLS] token of each view in the last transformer layer to a MLP to predict pose, and supervise pose with a quaternion loss and a translation loss.
Although this model can predict plausible reconstructions and poses, its performance is far worse than our full model where we use a differentiable PnP solver to predict poses.
We argue that this is because pose prediction has multiple local minimas, and the regression-based pose loss is more prone to such local minimas, compared with the EPro-PnP solver~\citep{chen2022epro} we use in this work. 
`MLP pose (Patch tokens)' take concatenated patch-wise features to a MLP for predicting pose. It aims to leverage more dense patch token information in the pose prediction.  The performance of this variant is roughly the same as `MLP pose (CLS token)'. 
`non-diff. PnP' removes the differentiable PnP prediction and only use the losses  $\Loss_\ImgPoint$ and $\Loss_\Imgalpha$. 
This way, we have a set of 3D-2D correspondences weighted by predicted opacity that are passed to a PnP solver for getting the poses. We find that this variant leads to worse performance than its differentiable PnP counterpart, due to the lack of learning proper confidence of 3D-2D correspondences. Note that PSNR$_{g.t.}$ reflects how well renderings of our predicted NeRF using ground-truth input poses match the input images, while PSNR$_{pred}$ measures how well renderings of our predicted NeRF using our pose predictions match the inputs. We observe that worse pose predictions tend to lead to worse reconstruction quality, as shown by the positive correlation between pose accuracy and PSNR$_{g.t.}$ scores in Tab.~\ref{tab:ablation_pose}.

\subsection{Additional Implementation Details} \label{sec:additional_details}

\begin{table}[h]
\centering
\begin{tabular}{l|l|l|l}
\hline
\multicolumn{2}{l|}{}             & Ours (S)     & Ours (L)    \\
\hline
\multirow{4}{*}{DINO Encoder}   & Image resolution     & 256$\times$256        & 512$\times$512      \\
                           & Patch size      & 16        & 16        \\
& Att. Layers      & 12        & 12      \\
                           & Attention channels       & 768       & 768      \\
                           & View encoding       & 768       & 768      \\
                           & Intrinsics-cond. MLP layers      & 5       & 5      \\
                           & Intrinsics-cond. MLP width      &  768      & 768     \\
                           & Intrinsics-cond. MLP act.       & GeLU       & GeLU      \\
\hline
\multirow{6}{*}{Transformer}   & Triplane tokens & $32\times32\times3$  & $32\times32\times3$ \\
                           & Attention channels        & 1024           & 1024       \\
                           & Attention heads  & 16 & 16 \\
                           & Attention layers     & 24 & 36 \\
 & Triplane upsample & 1         & 2        \\
                           & Triplane shape  &  $32\times32\times3\times32$             & $64\times64\times3\times32$\\
\hline
\multirow{5}{*}{Renderer} 
                           & Rendering patch size     & 64        & 128      \\
                           & Ray-marching steps      & 64        & 128       \\
                           & MLP layers     & 5        & 5       \\
                           & MLP width      & 64        & 64       \\
                           & Activation     & ReLU      & ReLU     \\
\hline
\multirow{3}{*}{Point MLP} 
                           & MLP layers     & 4        & 4      \\
                           & MLP width      & 512      & 512       \\
                           & Activation     & GeLU      & GeLU     \\
\hline
\multirow{6}{*}{Traininig} & Learning rate              &    4e-4       &       4e-4   \\
                           & Optimizer            &    AdamW       &  AdamW        \\
                           & Betas & (0.9, 0.95) &  (0.9, 0.95) \\
                           & Warm-up steps      &      3000     &  3000        \\
                           & Batch size per GPU & 16 & 8 \\ 
                           & \#GPUS & 8 & 128 \\
\hline
\end{tabular}
\caption{Configuration of our models. }
\label{tab:model_cofiguration} 
\end{table}

Our model uses a pre-trained DINO ViT as our image encoder.
We bilinearly interpolate the original positional embedding to the desired image size. For each view, its view encoding vector and camera intrinsics are first mapped to a modulation feature, and then passed to the adaptive layer norm block~\citep{hong2023lrm, Peebles2022DiT, huang2017arbitrary} to predict scale and bias for modulating the intermediate feature activations inside each transformer block (self-attention + MLP) of the DINO ViT~\citep{caron2021emerging}. Take the reference view as an example; its modulation feature $\mathbf{m}_r$ is defined as:
\begin{align}
\mathbf{m}_r = \mathrm{MLP^{intrin.}}([f_x, f_y, c_x, c_y]) + \VEncoding_r,
\end{align}
where $f_x, f_y, c_x, c_y$ are camera intrinsics, and $\VEncoding_r$ is the view encoding vector. We then use the modulation feature $\mathbf{m}_r$ in the same way as the camera feature in LRM~\citep{li2023instant3d}.

We then concatenate the image tokens with the learnable triplane position embedding to get a long token sequence, which is used as input to the single-stream transformer.
We use the multi-head attention with head dimension 64.
During rendering, the three planes are queried independently and the three features are concatenated as input of the NeRF MLP to get the RGB color and NeRF density. 
For per-view geometry prediction used for PnP solver, we use the image tokens output by the transformer with MLP layers to get the point predictions, the confidence predictions, and also the alpha predictions.

In our experiments we have models with two different sizes.
In the ablation studies as described in Sec.~\ref{sec:ablation}, the `Ours (S)' model has 24 self-attention layers, while the `Ours (L)` model has 36 self-attention leyers.
More details of the two model configurations are presented in Tab.~\ref{tab:model_cofiguration}.

We use the following techniques to save the GPU memory for our model training: 1) Mixed precision with BFloat16, 2) deferred back-propagation in NeRF rendering ~\citep{zhang2022arf}, and 3) Gradient checkpointing at every 4 self-attention layers.
We also adopt the FlashAttention V2~\citep{dao2023flashattention} to reduce the overall training time.

\subsection{Additional results}
{\bf Category-level results on CO3D dataset.} In Tab.~\ref{tab:pose_compare_co3d} we report the per-category results and comparisons to RelPose++ on held-out CO3D test set provided by RelPose++~\cite{lin2023relposepp}. We outperform RelPose++ (\textit{w/ bg}) on 8 out of 10 categories, despite that we are not trained on CO3D training set while RelPose++ is. In addition, our model is now limited to handle images without background; hence we use the masks included in the CO3D dataset to remove background before testing our model. The masks, however, seem to be very noisy upon our manual inspection; this negatively influenced our model's performance, but not RelPose++ (\textit{w/ bg}). An interesting future direction is to extend our model to support images with background to in order to lift the impacts of 2D mask errors.

\begin{table}[htb]
    \caption[]{Evaluation results on GSO data~\citep{gso} rendered by FORGE~\citep{jiang2022forge}. We note that these renderings are a bit darker than majority of our training images, but our model still generalizes well to this dataset. Our model produces sharper renderings than FORGE (indicated by the higher SSIM score), while producing more accurate camera estimates. }
\resizebox{\linewidth}{!}{
    \centering
    \begin{tabular}{c|c|c|c|c|c|c|c|c}
    \hline
    Method & R. error & Acc.@$15^{\circ}$ & Acc.@$30^{\circ}$ & T. error & PSNR$_{\rm g.t.}$ & PSNR$_{\rm pred.}$ & SSIM$_{\rm g.t.}$ & SSIM$_{\rm pred.}$  \\
    \hline
    {\small FORGE } & 50.20 & 0.253 & 0.514	& 0.573 & 21.25 & 22.90 & 0.767 & 0.793 \\
    {\small FORGE (refine) } & 49.02 & 0.307 & 0.527 & 0.548 & 22.08 & \textbf{25.89} & 0.767 & 0.838  \\
    {\small Ours} & \textbf{8.37} & \textbf{0.908} & \textbf{0.954} & \textbf{0.105} & \textbf{23.05} & 24.42 & \textbf{0.860} & \textbf{0.886} \\
    \hline
    \end{tabular}
}    
    \label{tab:pose_compare_gso_dark}
\end{table}

\begin{table}[htb]
    \caption[]{Ablation study of different pose prediction methods on the GSO data~\citep{gso}. Ablations are conducted using methods are our small model, i.e., `Ours (S)'. Compared with our method of predicting per-view coarse geometry followed by differentiable PnP~\citep{chen2022epro}, the MLP-based pose prediction method conditioning on either the per-view CLS token or the concatenated patch tokens perform much worse due to the lack of explicit geometric inductive bias (either 3D-2D correspondences or 2D-2D correspondences) in pose registrations. Besides, we also find that differentiable PnP learns to weigh the 3D-2D correspondences induced from the per-view predicted coarse geometry properly, resulting a boost in pose estimation accuracy. }
    \centering
    \begin{tabular}{c|c|c|c|c|c|c}
    \hline
    Setting & R. error & Acc.@$15^{\circ}$ & Acc.@$30^{\circ}$ & T. error & PSNR$_{\rm g.t.}$ & PSNR$_{\rm pred.}$\\
    \hline
    {\small diff. PnP (our default setting)} & \textbf{13.08} & \textbf{0.848} & \textbf{0.916} & \textbf{0.135} & \textbf{23.80} & \textbf{22.82} \\
    {\small MLP pose (CLS token)} & 25.32 & 0.655 & 0.809 & 0.264 & 22.27 & 19.80 \\
    {\small MLP pose (Patch tokens)} & 21.60 & 0.688 & 0.836 & 0.230 & 22.02 & 19.76 \\
    {\small non-diff. PnP} & 22.03 & 0.570 & 0.814 & 0.236 & 23.56 & 18.65  \\
    \hline
    \end{tabular}
    \label{tab:ablation_pose}
\end{table}

\subsection{Additional cross-dataset evaluations}

To further demonstrate the generalization capability of our model, we evaluate our model (trained on a mixture of Objaverse and MVImgNet)
on another version of GSO dataset~\citep{gso} (which is rendered by the FORGE paper). Note that these renderings are a bit darker than majority of our training images, but as shown in Tab.~\ref{tab:pose_compare_large}, our model still generalizes well to this dataset. Our model produces sharper renderings than FORGE with and without its per-scene optimization-based refinement (indicated by the higher SSIM score), while producing much more accurate camera estimates.  Note that PSNR$_{g.t.}$, SSIM$_{g.t.}$ reflect how well renderings of our predicted NeRF using ground-truth input poses match the input images, while  PSNR$_{pred}$,  SSIM$_{pred}$  measures how well renderings of our predicted NeRF using ground-truth input poses match the inputs.

\subsection{Scaling up training of RelPose++}
\label{sec:scale_up_relposepp}
To further demonstrate our method's superiority over the baseline method RelPose++~\citep{lin2023relposepp}, 
we re-train RelPose++ on the Objaverse dataset until full convergence for a more fair comparison. We then compare the re-trained model with our model (`Ours (S)' and `Ours (L)') trained on exactly the same Objaverse renderings in Tab.~\ref{tab:revision_pose_compare_all}. The re-trained RelPose++ using Objaverse does improve over the pretrained one using CO3D on the unseen test sets, OmniObject3D, GSO and ABO. However, our models (both `Ours (S)' and `Ours (L)') consistently outperform the re-trained baseline by a large margin in terms of rotation and translation prediction accuracy. We attribute this to our joint prediction of NeRF and poses that effectively exploit the synergy between these two tasks; in addition, unlike RelPose++ that regresses poses, we predict per-view coarse point cloud (supervised by distilling our predicted NeRF geometry in an online manner) and use a differentiable solver to get poses. This make us less prone to getting stuck in pose prediction local minimas than regression-based predictors, as also pointed out by~\cite{chen2022epro}.

\begin{table}[htb]
    \centering
    \caption[]{Comparisons of cross-dataset generalization on GSO~\citep{gso}, ABO~\citep{abo}, OmniObject3D~\citep{wu2023omniobject3d} with RelPose++~\citep{lin2023relposepp} using the author-provided checkpoint (trained on CO3D~\citep{reizenstein21co3d} and our \textbf{re-trained} checkpoint (trained on Objaverse~\citep{objaverse}). `Ours (S)' and `Ours (L)' are trained only on Objaverse as well for fair comparison. Though the \textbf{re-trained} RelPose++ improves over the pretrained version, we (both `Ours (S)' and `Ours (L)') still achieve much better pose prediction accuracy than it.  }
    \begin{tabular}{c|c|c|c|c}
    \hline
    \multicolumn{5}{c}{\bf OmniObject3D} \\
    \hline
     Method & R. error $\downarrow$ & Acc.@$15^{\circ}$ $\uparrow$ & Acc.@$30^{\circ}$ $\uparrow$ & T. error $\downarrow$ \\
    \hline
    {\small RelPose++ (w/o bg, pretrained)} & 69.22 & 0.070 & 0.273 & 0.712 \\
    {\small RelPose++ (w/o bg, Objaverse)} & 58.67 & 0.304 & 0.482 & 0.556 \\
    {\small Ours (S)} & 15.06 & 0.695 & 0.910 & 0.162 \\
    {\small Ours (L)} & \textbf{7.25} & \textbf{0.958} & \textbf{0.976} & \textbf{0.075} \\
    \hline
    \multicolumn{5}{c}{\bf GSO} \\
    \hline
    Method & R. error $\downarrow$ & Acc.@$15^{\circ}$ $\uparrow$ & Acc.@$30^{\circ}$ $\uparrow$ & T. error $\downarrow$\\
    \hline
    {\small RelPose++ (w/o bg, pretrained)} & 107.49 & 0.037 & 0.098 & 1.143 \\
    {\small RelPose++ (w/o bg, Objaverse)} & 45.58 & 0.600 & 0.686 & 0.407  \\
    {\small Ours (S)} & 13.08 & 0.848 & 0.916 & 0.135  \\
    {\small Ours (L)} & \textbf{2.46} & \textbf{0.976} & \textbf{0.985} & \textbf{0.026} \\
    \hline
    \multicolumn{5}{c}{\bf ABO} \\
    \hline
    Method & R. error $\downarrow$ & Acc.@$15^{\circ}$ $\uparrow$ & Acc.@$30^{\circ}$ $\uparrow$ & T. error $\downarrow$ \\
    \hline
    {\small RelPose++ (w/o bg, pretrained)} & 102.30 & 0.060 & 0.144 & 1.103 \\
    {\small RelPose++ (w/o bg, Objaverse)} & 45.39 & 0.693 & 0.708 & 0.395 \\
    {\small Ours (S)} & 26.31 & 0.785 & 0.822 & 0.249  \\
    {\small Ours (L)} & \textbf{13.99} & \textbf{0.883} & \textbf{0.892} & \textbf{0.131} \\
    \hline
    \end{tabular}
    \label{tab:revision_pose_compare_all}
\end{table}

\end{document}